\documentclass{article}

\usepackage{natbib}

\usepackage{PRIMEarxiv}

\usepackage[utf8]{inputenc} % allow utf-8 input
\usepackage[T1]{fontenc}    % use 8-bit T1 fonts
\usepackage{hyperref}       % hyperlinks
\usepackage{url}            % simple URL typesetting
\usepackage{booktabs}       % professional-quality tables
\usepackage{amsfonts}       % blackboard math symbols
\usepackage{nicefrac}       % compact symbols for 1/2, etc.
\usepackage{microtype}      % microtypography
\usepackage{lipsum}
\usepackage{graphicx}
\usepackage{svg}
\usepackage{amsmath}
\graphicspath{ {./images/} }
\usepackage[justification=centering]{caption}
\usepackage[english]{babel}
\usepackage{wrapfig}
\usepackage{caption}
\usepackage{subcaption}
\usepackage{chngcntr}
\usepackage{multicol}
\usepackage{caption} 
\hypersetup{hidelinks}

\usepackage{csquotes}

\title{Machine Learning based medical image deepfake detection: A comparative study}

\author{
 Siddharth Solaiyappan \\
  Northwood High School\\
  Irvine Unified School District\\
 Irvine, CA 92620 \\
  \texttt{siddharthsolaiyappan@gmail.com} \\
   \And
 Yuxin Wen  \\
 Fowler School of Engineering\\
 Chapman University\\
 Orange, CA 92866 \\
 \texttt{yuwen@chapman.edu} \\
}

\begin{document}

\maketitle{}

%\newpage

\begin{abstract}
Deep generative networks in recent years have reinforced the need for caution while consuming various modalities of digital information. One avenue of deepfake creation is aligned with injection and removal of tumors from medical scans. Failure to detect  medical deepfakes can lead to large setbacks on hospital resources or even loss of life. This paper attempts to address the detection of such attacks with a structured case study. Specifically, we evaluate eight different machine learning algorithms, which including three conventional machine learning methods, support vector machine, random forest, decision tree, and five deep learning models, DenseNet121, DenseNet201, ResNet50, ResNet101, VGG19,  on distinguishing between tampered and untampered images.For deep learning models, the five models are used for feature extraction, then fine-tune for each  pre-trained model is performed.  The findings of this work show near perfect accuracy in detecting instances of tumor injections and removals. 


\end{abstract}

% keywords can be removed

\section{Introduction}
{
The use of medical imaging techniques, including ultrasound, computed tomography (CT), magnetic resonance imaging (MRI) and X-ray scans, has been playing essential role in diagnosis and treatment of various diseases due to their ability to view the interior of a body in a non-invasive manner. Compared to conventional 2D X-ray images, CT scans provide  3D, cross-sectional images of the body, allowing for views of the bone, soft tissue and blood vessels from multiple angles at the same time \citep{ct}.} CT scans are usually stored as Digital Imaging and Communications in Medicine (DICOM) format in a Picture and Archive Communication System (PACS). PACS servers allow ease of transfer of medical imagery between hospital scanners, radiologists workstations, and network intermediates. Medical centers are often equipped with outdated security measures and legacy software, allowing for open access to data. For example, researchers from McAfee equipped with 3D printers reconstructed a model of a pelvis from unsecured volumetric DICOM data \citep{mcafee}. Deceptive attacks of added tumors on the original image can cause patients to receive unnecessary treatment, and cost millions of dollars in hospital resources. Removing a tumor will deprive the patient of needed treatment and severely advance an existing condition, or even lead to loss of life. 80 million CT scans are conducted in the U.S. every year \citep{nctscans}. With this frequency of medical imaging, such attacks can be politically motivated, financially motivated with insurance fraud, or other severe financial or life loss. Therefore, there is a pressing need to develop effective and reliable methods for the  detection and identification of image tampering. 

Copy-move and image-splicing are two commonly used image tampering methods. Copy-move involves duplicating an area of non-interest over the target region, masking the region of interest from the viewer. This method can also be used to duplicate the target region and increase the frequency of regions of interest.  Image-splicing  follows a similar procedure to copy-move, difference being, the duplicated region of interest in image-splicing originates from an external image. With the advancement of  machine learning and deep learning techniques, attackers now are capable of learning from collections of images to create an image that looks authentic to human viewers, by using generative adversarial networks (GANs).  GANs, introduced by  \citet{goodfellow2014generative}, consist of two neural networks which work
against each other: the generator and the discriminator. The
generator creates fake samples with the aim of deceiving the
discriminator, and the discriminator learns to differentiate
between real and fake samples. If applied to image tampering, the content of a 3D medical image can be easily altered maliciously in a realistic and automated way. \citet{DBLP:journals/corr/abs-1901-03597} developed a CT-GAN framework, which is capable of adding or removing cancer tumors from a patient's CT lung scans and can be executed by a malware autonomously. 
To evaluate the extent to which the CT-GAN attack can fool expert radiologists and state-of-the-art AI, in the case study, Mirsky et al. selected 3 radiologists, each with 2, 5, and 7 years of experience respectively, and a trained lung cancer AI screening model,  which won the 2017 Kaggle Data Science Bowl (a 1 million competition for diagnosing lung cancer) \citep{DBLP:journals/corr/abs-1711-08324}, to diagnose a batch of scans with, and without tampering. This experiment was conducted through 2 distinct trials: blind and open. In the blind trial, the scan samples were provided to the radiologists with no knowledge of tampering. The results of the blind trial showed 99\% of tumor injected scans were misdiagnosed as malignant, and 94\% of tumor removed scans were misdiagnosed as benign. Once made aware of the tampering mechanisms in an open trial, the misdiagnosis rates decreased to 60\% and 87\% respectively. In addition, they found the AI lung cancer screening model misdiagnosed 100\% of the tampered patients in the blind trial. These results go to show that current medical systems have become increasingly vulnerable to the image tampering attack.

{In response to these image tampering, various methods have been proposed to detect and localize tampered images. These techniques can be  split into two types, namely, active and passive detection  \citep{birajdar2013digital}. In active detection, such as digital watermarking and digital signature,   a known authentication code is embedded into the image content before the images are sent through an unreliable public channel. 
The data is authenticated by comparing the embedded code with its original instance. \citet{9682744} proposed a region-based hybrid medical image watermarking scheme using hybrid domain watermark techniques for higher imperceptibility, security, tamper detection accuracy, and authentication, even under unintentional attacks. \citet{Pravin_Savaridass_2021} and {\citet{Mohammed_2021}}, implemented a hybrid watermarking technique by the combination of discrete Wavelet transform (DVT) and singular value decomposition. In their work, the images have been tested against attacks like salt and pepper noise, Gaussian noise, and filtering attack. Despite the noticeable advantages, active detection requires special hardware or software to embed the authentication code within the image before it is distributed. On the contrary, passive detection has gained popularity for its capability to be implemented without any prerequisite information about the image. passive detection is performed by identifying the changes in local features and the entire image by comparing the frequency domain properties, or statistical information of the image.  \citet { thakur2018blind} used  DVT  to decompose the image into sub images and obtain coefficient for each sub image, then used speed-up robust features methods  to extract features and employed Support Vector Machine (SVM)  to perform classification for splicing forgery detection.  In passive detection, the accuracy highly depends on the extracted features. However, it is difficult  to identify which feature should be extracted for tampering detection. }

{Recently, there has been a research trend of switching the image tampering detection mechanism from conventional hand-crafted feature extraction based machine learning models to  deep learning frameworks. The emergence of deep learning has proved very powerful at distilling complex hidden features in the data and, thus, these methods typically demonstrate good performance. The main advantage of deep learning approaches lies in the capability  of automated extraction of complex data representations through end-to-end training from raw data, which significantly reduces the effort of manual feature engineering.  \citet{kadam2022efficient} presented a mask R-CNN with MobileNet framework  to detect and identify copy-move and image-splicing forgeries. \citet{krishnaraj2022design} provided an automated deep learning-based fusion model for image tampering detection by combining models of generative adversarial networks (GANs) and densely connected networks (DenseNets). \citet{koul2022efficient} used convolutional neural network (CNN) for image forgery detection. We noticed that all of these deep learning methods are designed to detect specific forms of tampering, such as copy-move, image-splicing.  Little work has been done  to explicitly address the design of robust forgery detection  generated by GAN. To mitigate this gap,  we explore multiple machine learning approaches to detect image tampering generated by the CT-GAN framework \citep{DBLP:journals/corr/abs-1901-03597}. We evaluate eight different machine learning algorithms, which include three conventional machine learning methods, i.e., support vector machine (SVM), random forest, decision tree, and five deep learning models, DenseNet121, DenseNet201, ResNet50, ResNet101, VGG19, to distinguish between tampered and untampered images. For deep learning models, the five models are used for feature extraction, then fine-tuned.  }

The rest of this paper is organized as follows. Section 2  describes the dataset and machine learning models we used for comparative analysis, Section 3 demonstrates the classification results and discussion about the findings. Section 4 concludes the paper with a discussion of potential future work.

 \section{Materials and Methods}
In our case study, all the untampered data  is from  the largest publicly available annotated CT database, LIDC-IDRI  \citep{lidcdata}, and  the tampered data is from CT-GAN dataset  (\cite{DBLP:journals/corr/abs-1901-03597}; \cite{UCImlrepo}). All scans are patient records in DICOM format, and are separated by uniquely labeled folders that correspond to each patient ID.

The combined dataset is grouped into three categories:

\begin{itemize}
% \item True Benign (TB) - Natural scan without cancerous growth
% \item True Malign (TM) - Natural scan with cancerous growth
\item Untampered - Natural scans with or without cancerous growth
\item False Benign (FB) - Scan with artificially removed cancerous growth
\item False Malignant (FM) - Scan with artificially injected cancerous growth
\end{itemize}{}

% \begin{itemize}
% \item 3.1.1, 3.1.2 - Train: 187 ; Test: 27 
% \item 3.1.3, 3.2.2 - Train: 9112 ; Test: 1880 
% \item 3.2.1 - Train: 193 ; Test: 32 
% \item 3.3 - Train: 6834 ; Test: 1407 
% \end{itemize}{}

{To investigate the capabilities of machine learning algorithms for image tampering detection, eight different machine learning algorithms, which include three conventional machine learning methods, SVM, random forest, decision tree, and five deep learning models, DenseNet121, DenseNet201, ResNet50, ResNet101, VGG19, are applied to distinguish between tampered and untampered images. SVM operate by finding  a separating hyperplane with the largest separation, or margin, between the two classes. The decision boundary can be found by minimizing an optimization problem \citep{chang2011libsvm}. The key idea of Decision Tree is to divide the dataset into smaller subsets in the form of nodes and branches. Random Forest is a variant of  Decision Tree. Random Forest creates a set of  Decision Trees using random resampling on the training set. For classification tasks, each Decision Tree  votes for a particular target class and the class with the majority vote is the outputs of the Random Forests. VGG19 is a CNN that is 19 layers deep. VGG, is a popular neural network architecture proposed by Karen Simonyan and Andrew Zisserman from the University of Oxford \citep{simonyan2014very}. DenseNet (Dense Convolutional Network) \citet{huang2017densely} is an architecture that focuses on making the deep learning networks go  deeper, but at the same time making them more efficient to train, by using shorter connections between the layers. The  idea of ResNet is introduced a so-called “identity shortcut connection” that skips one or more layers to address the vanishing gradient issue and  mitigate the degradation (accuracy saturation) problem \citep{he2016deep}. The Densenet-201 is larger at over 77MB in size , in comparison to the roughly 31 MB size of DenseNet121.} All deep learning neural networks follow the same architecture of a convolution base, a dense 10-neuron layer with rectified linear activation (ReLU), and a softmax activated 2-neuron layer. All pretrained convolution layers were frozen while other layers were allowed to train.In all case studies, the train-test split ratio is 85\%:15\%. Table ~\ref{table:numbers}  lists the number of images in each category we used.

\begin{table}[!htp]\centering
\caption{Number of  images by type} %\label{tab: }
\begin{tabular}{lrrrr}\toprule
Untampered &FB &FM &Total \\\midrule
2278 &65 &41 &2384 \\
\bottomrule
\end{tabular}
\label{table:numbers}
\end{table}

{In the experiment, to select the best model for each classifier, five-fold cross-validation is used on the training data set. Specifically, the training data is split into 5 groups. For each iteration, 1 group is reserved as the validation data, and all the rest of the groups are used for training. Then the validation data is fed into the trained classifier. This process is repeated 5 times. To decide optimal hyperparameters, grid search is applied when performing five-fold cross-validation for each classifier, which is a tuning technique that exhaustively generates candidates from a grid of parameter values, then builds a model for every combination of hyperparameters specified and evaluates each model.  }

In order to transition from three dimensional voxel data to CNN compatible two dimensional image data, we first iterate through the list of training set patient numbers and find all corresponding (z,x,y) coordinates for tampered regions per patient. The z.dcm file is then read and the pixel array of the DICOM file is outputted and saved in .jpg format as a single channel gray image. This is done multiple times on most occasions as a single voxel often has multiple 2D slices where tampering has occurred. Due to the barrier of memory intensivity, the possibility of extracting a 3D sample around the tumor site was not implemented. Conventional ML models were fitted with the raw pixel array without any feature extraction, whereas the deep networks had data generator classes where the respective network’s preprocessing input was applied.

\section{Experiments and Results}

This section evaluates the classification capabilities of eight different classifiers to distinguish tampered and untampered CT scans. {To conduct the comparison, the code is written in Python, conventional machine learning computations are performed on a 2.3 GHz Intel Xeon CPU, and deep learning models are trained using a NVIDIA Tesla P100-PCIE GPU.} Firstly, in Section 3.1, we compare the performance analysis for FM image detection using raw, localized, and augmented images. In Section 3.2, FB image detection is performed using delocalized, and augmented images. Finally, in Section 3.3,   multi-class classification, i.e., FB, FM, untampered, for delocalized data augmented images is explored.

% \pagebreak
% \newpage

\subsection{False Malign detection}
The primary focus of this section is to detect false malignant from untampered scans. In the blind trial conducted by the authors of CT-GAN (\cite{DBLP:journals/corr/abs-1901-03597}), 3 radiologists and an AI were asked to classify 80 scans as either benign or malignant. The injected tumor misdiagnosis rate for radiologists and AI was 99.2\% and 100\% respectively. The AI model used was a 3D Deep Leaky Noisy-or Network (\cite{DBLP:journals/corr/abs-1711-08324}) trained on the Lung Nodule Analysis (LUNA) dataset (\cite{LUNA_DBLP:journals/corr/SetioTBBBC0DFGG16}).

\subsubsection{Raw Image Classification}
\begin{figure}[h] % picture
    \centering
    \includegraphics[scale=0.35]{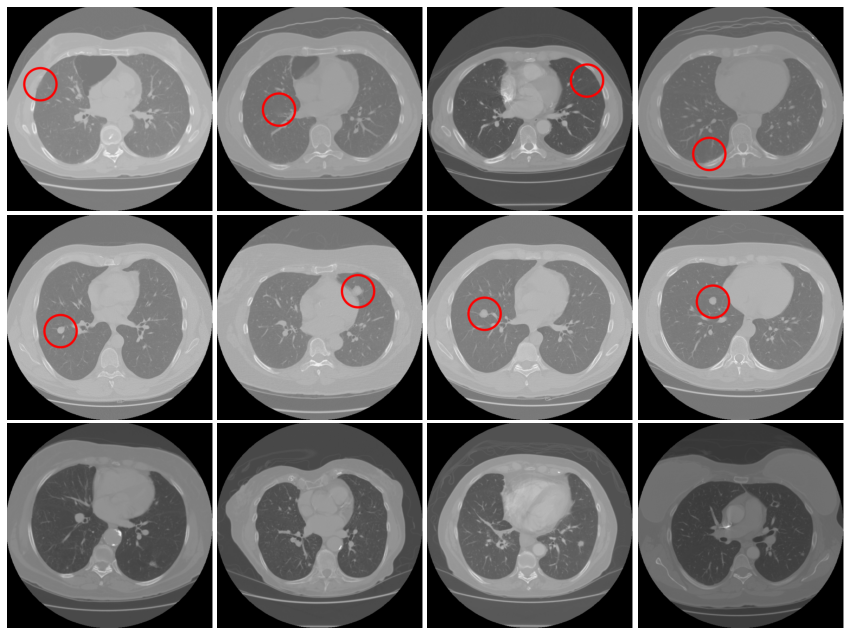}
    \caption{Raw CT scan images with tampered regions circled}  
    % \paragraph{}
    \caption*{Row 1: FB samples, Row 2: FM samples,  Row 3: Untampered samples}
    \label{figure:figure1}
\end{figure}

In the paper CT-GAN (\cite{DBLP:journals/corr/abs-1901-03597}) and (\cite{nodulereport}), Authors have provided annotations to voxel coordinates of tumor sites and nodule locations in the tampered and untampered datasets respectively. Figure  ~\ref{figure:figure1} shows annotated 2D slices extracted from the corresponding 3D CT scans. Tampered regions are circled as red. We first conduct the classification using raw images.

\begin{figure}[h] % picture
    \centering
    \includegraphics[scale=0.45]{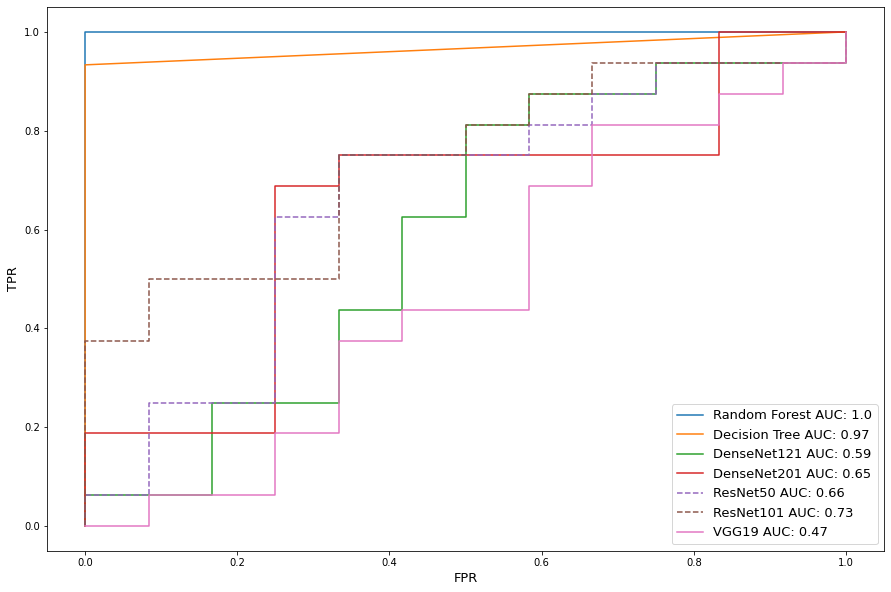}
    \caption{Classifier ROC curves with raw images}
    \label{figure:figure2}
\end{figure}

The eight models are trained without data preprocessing on all tampered images, and a subset of untampered images to balance data quantity for classes. Table ~\ref{table:table2} shows train and test performance metrics for the classifiers. The SVM, Random forest and Decision tree perform much better then deep learning models. Insufficient data severely hindered deep learning models' accuracy due to overfiting issue.
Figure ~\ref{figure:figure2} shows the receiver operator characteristic (ROC) curves, where TPR and FPR represent true positive rate and false positive rate respectively. {The ROC curve is created by plotting the TPR against the FPR at various threshold settings, which is a commonly used plot to qualify the diagnostic ability of a classifier. The area under the curve (AUC) summarizes the quality of classification and is used as another measure of accuracy, where an AUC of 0.5 indicates a random classifier with no predictive credibility.} The results show ResNet101 is able to reach 80\% TPR with the hindrance of a 50\% FPR rate and would be considered poor (\cite{el2009relationship}). FPR addresses classification accuracy of tampered scans and is required to be minimized as much as possible in order to avoid life threatening misclassification. Given low quantity and high dimensional data, the bootstrap aggregation and random feature selection allow the random forest to produce a  high performance. However with limited test sample presence, further testing is required to determine true accuracy levels with features at scale.

\begin{figure}[h] % picture
    \centering
    \includegraphics{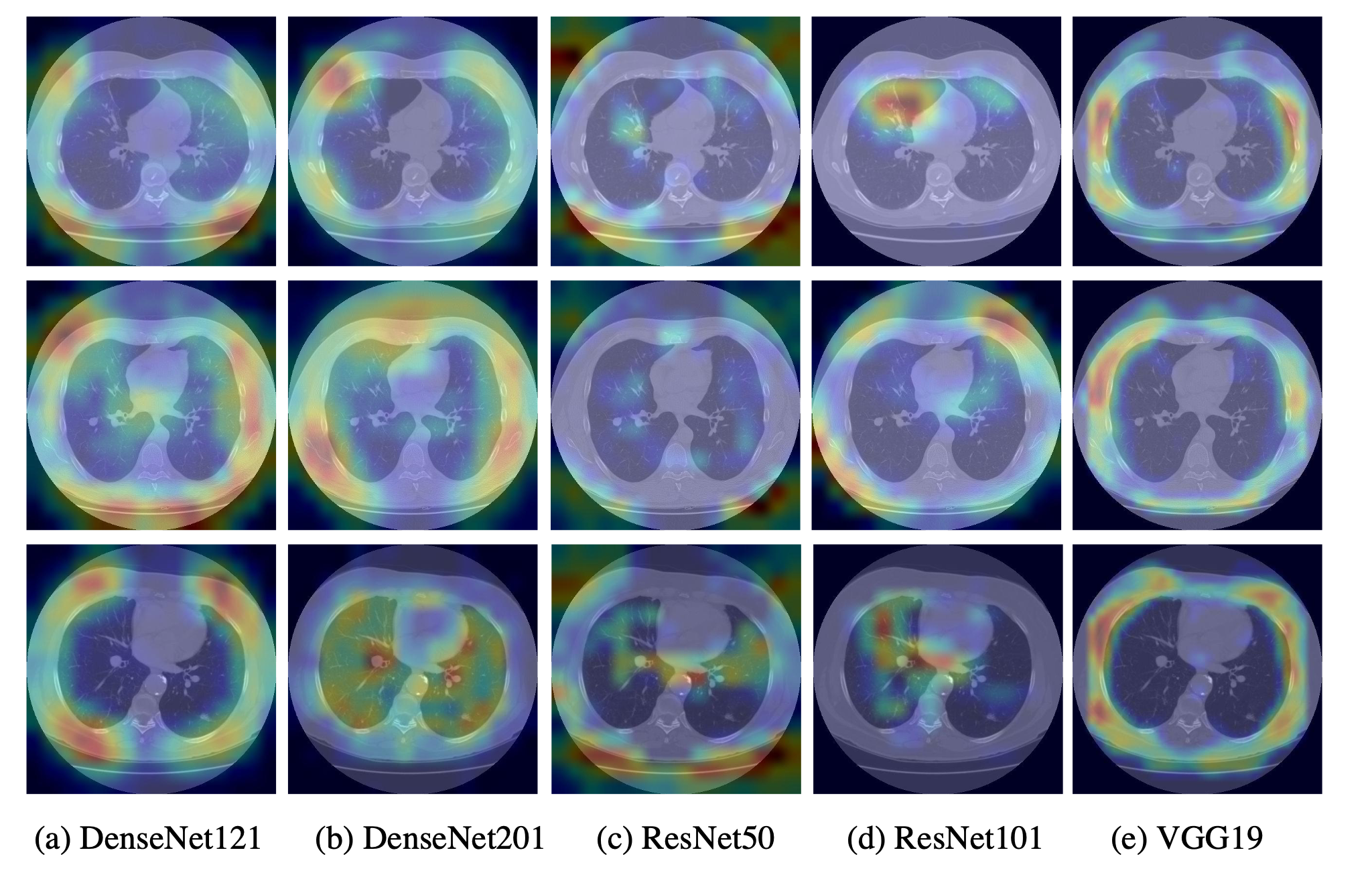}
    \caption{Grad-CAM visualization for deep learning models}
    \label{figure:figure3}
\end{figure}

{Gradient-weighted class activation mapping (Grad-CAM) is applied on deep learning models to clearly interpret the prediction results and provide an explainable visual \citep{gite2022enhanced}. Grad-CAM, introduced by \citet{selvaraju2017grad}, uses the gradient of any target concept and places it in the final convolution layer to generate a coarse localization map that highlights the key regions in the image to predict the required concepts, which can be used to help users establish appropriate trust in predictions based on deep networks. Figure ~\ref{figure:figure3} shows the Grad-CAM visualization applied to the first column of Figure ~\ref{figure:figure1} using 5 deep learning models. The overlaid heat-maps indicate increasing model prediction confidence from 0.0 to 1.0. As we can see, the regions of abnormalities are localized in some models. But  irrelevant regions are also included by almost all models, which lead to unsatisfactory results.}

% \pagebreak

\begin{table}[!htp]\centering
\caption{Classifier performance using raw images}%\label{tab: }
% \scriptsize
\begin{tabular}{lrrrrr}\toprule
Classifier &Train Accuracy &Train Loss &Test Accuracy &Test Loss \\\midrule
SVM &0.98 &- &\textbf{0.95} &- \\
Random Forest &\textbf{1} &- &\textbf{0.95} &- \\
Decision Tree &0.96 &- &\textbf{0.95} &- \\
DenseNet121 &0.82 &0.38 &0.54 &1.58 \\
DenseNet201 &0.86 &0.34 &0.57 &1.27 \\
ResNet50 &0.99 &\textbf{0.08} &0.56 &\textbf{0.58} \\
ResNet101 &0.98 &0.11 &0.61 &1.05 \\
VGG19 &0.98 &0.11 &0.43 &1.62 \\
\bottomrule
\end{tabular}
\label{table:table2}
\end{table}

% \pagebreak

\subsubsection{Tumor Site Localization}

\begin{figure}[h] % picture
    \centering
    \includegraphics[scale=.212]{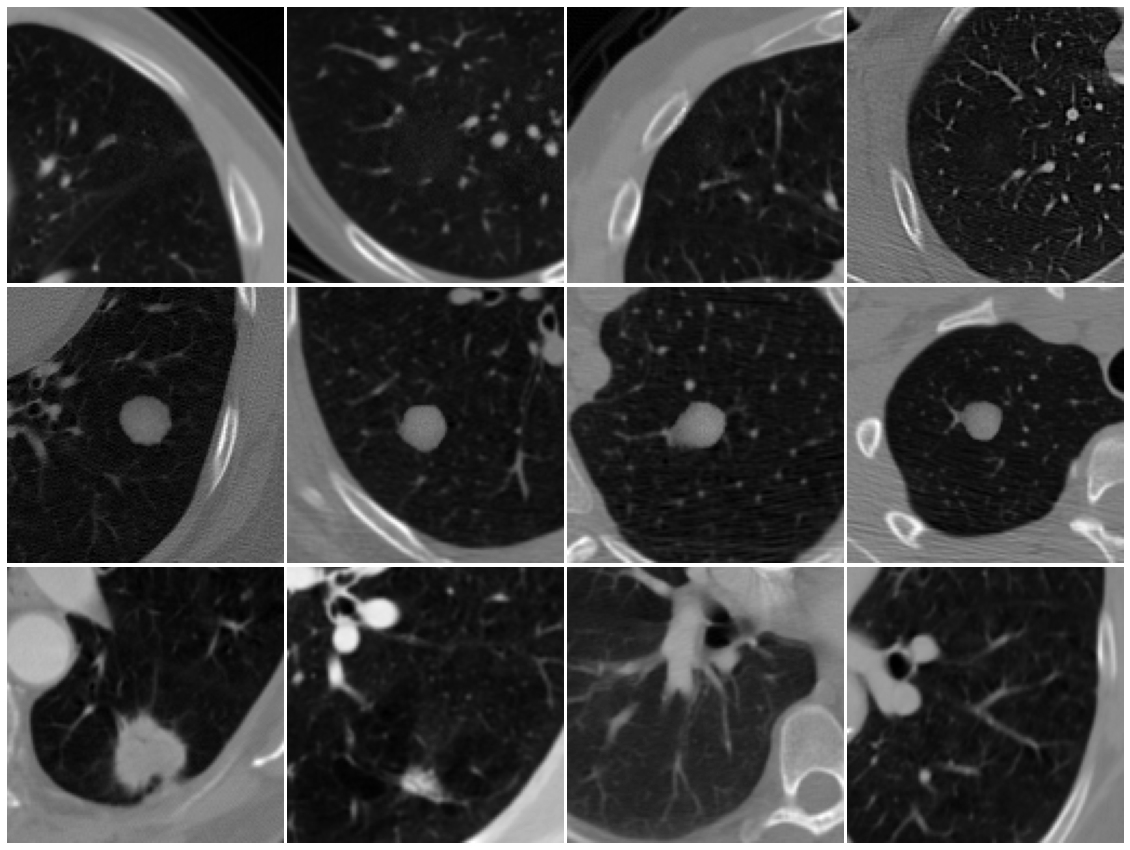}
    \caption{Examples of localized images}
    % \paragraph{}
    \caption*{Row 1: FB samples, Row 2: FM samples,  Row 3: Untampered samples}
    \label{figure:figure4}
\end{figure}

\begin{figure}[h] % picture
    \centering
    \includegraphics[scale=0.45]{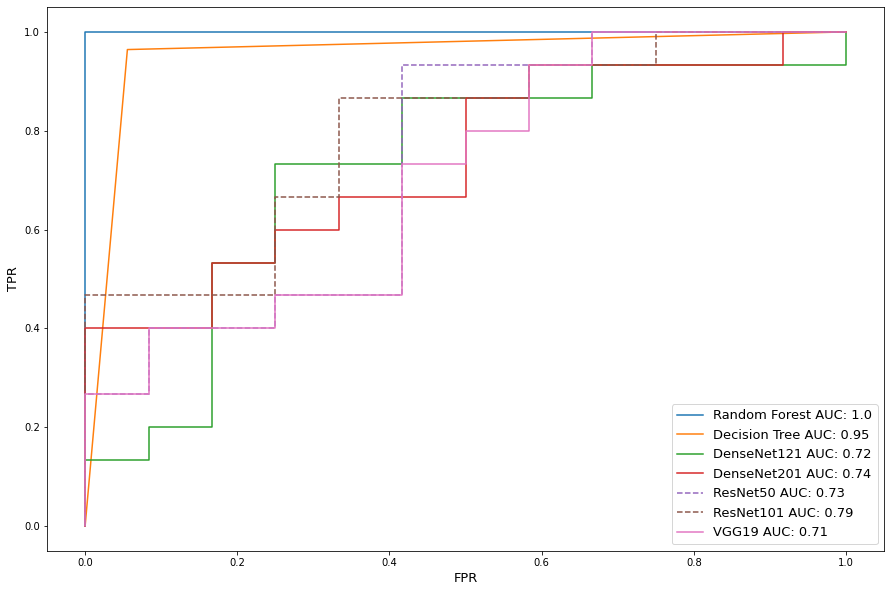}
    \caption{Localized image classifier ROC Curves}
    \label{figure:figure5}
\end{figure}

Data from the blind experiment showed radiologists can identify the locations of tumor sites. However, the challenge was distinguishing between natural and artificially generated tumors. We test localizing the tumor site in an attempt to mimic this difficulty. The annotated coordinate data from the dataset allows for tumor site localization: slicing a 128x128 figure from the generated pixel array using the tumor site as center. This section uses the same scans as in Section 3.1.1, but localizes the scans to create the new dataset. FB scans are still used for training as annotated locations of tumor sites are provided. However, practical use of the models are restricted to FM scan detection as FB site locations will be unknown to detection mechanisms and cannot be localized. Examples of localized images are shown in Figure ~\ref{figure:figure4}. Table ~\ref{table:table3} demonstrates the classification performance using localized images. We conclude that the region of interest (ROI) extraction increased classifier accuracy for all model types. Due to a decrease in number of learnable features, and reasons previously mentioned, random forest and SVM both reach perfect accuracy with small testing samples. In terms of deep learning, Figure ~\ref{figure:figure5} shows VGG19 can achieve a ~75\% true positive rate with a ~45\% false positive rate, and ResNet101 can achieve a ~90\% TPR with a ~35\% FPR dependence. Though deep learning accuracy increased, the scores indicate little deviation from guessing accuracy for a binary classifier.

\begin{table}[h]\centering
\caption{Localized image classifier performance}
\begin{tabular}{lrrrrr}\toprule
Classifier &Train Accuracy &Train Loss &Test Accuracy &Test Loss \\\midrule
SVM &0.97 &- &\textbf{1} &- \\
Random Forest &\textbf{1} &- &\textbf{1} &- \\
Decision Tree &\textbf{1} &- &0.98 &- \\
DenseNet121 &\textbf{1} &0.13 &0.63 &0.87 \\
DenseNet201 &\textbf{1} &\textbf{0.02} &0.64 &1.08 \\
ResNet50 &\textbf{1} &0.07 &0.67 &0.92 \\
ResNet101 &\textbf{1} &0.03 &0.63 &1 \\
VGG19 &0.79 &0.5 &0.71 &\textbf{0.78} \\
\bottomrule
\end{tabular}
\label{table:table3}
\end{table}

% \pagebreak
\newpage

% \newpage

% \begin{figure}[h]
%   \centering
%   \includesvg[scale=0.35]{roc128.svg}
%   \caption{Localized image classifier ROC Curves}
% \end{figure}

% \pagebreak
% \newpage
%  \clearpage

\subsubsection{Data Augmentation}

 The effects of data quality and quantity in medical applications have been studied (\cite{BARRAGANMONTERO202152}). Higher performance was established with increase in data quality, but negligible changes were detected with increase in dataset size. The possibility of generating high quality data via CT-GAN (\cite{DBLP:journals/corr/abs-1901-03597}) was deemed impractical due to the algorithm’s demanding hardware dependencies. As such, with the current objective of increasing specifically deep learning model accuracy, we re-conduct the localized tumor site test with data augmentation for both classes. The technique used closely resembles that used to augment and train CT-GAN (\cite{DBLP:journals/corr/abs-1901-03597}), but in one less dimension:
\begin{itemize}
\item Flip over x-axis, y-axis, and both axes
\item Combinations of x,y shifts; 4 units in specified direction
\item Rotation of 360 degrees in 6 degree increments
\end{itemize}{}

% \begin{figure}[h] % picture
%     \centering
%     \includegraphics[scale=1.0]{ CS 1 Augmented Samples.png}
%     \caption{Row 1: FB samples, Row 2: FM samples,  Row 3: Untampered samples}
% \end{figure}

% Classifier accuracy is as follows. More transfer learning models were included because of relative increase in performance from previous tests.

\begin{table}[!htp]\centering
\caption{Augmented localized image classifier performance}%\label{tab: }
\begin{tabular}{lrrrrr}\toprule
Classifier &Train Accuracy &Train Loss &Test Accuracy &Test Loss \\\midrule
Support Vector Machine &0.984 &- &0.978 &- \\
Random Forest &\textbf{1} &- &0.979 &- \\
Decision Tree &\textbf{1} &- &0.925 &- \\
DenseNet121 &0.995 &\textbf{0.03} &\textbf{0.991} &0.042 \\
DenseNet201 &0.991 &0.026 &0.987 &0.077 \\
ResNet50 &0.995 &\textbf{0.03} &\textbf{0.991} &\textbf{0.022} \\
ResNet101 &0.989 &0.029 &0.98 &0.06 \\
VGG19 &0.977 &0.995 &0.949 &0.547 \\
\bottomrule
\end{tabular}
\label{table:table4}
\end{table}

% \begin{figure}[h]
% \begin{subfigure}{.5\textwidth}
%   \centering
%   % include first image
%   \includegraphics[width=.8\linewidth]{ROC 1.3.1.png}  
%   \caption{DenseNet201 AUC score: 1.0}
%   \label{fig:sub-first}
% \end{subfigure}
% \begin{subfigure}{.5\textwidth}
%   \centering
%   % include second image
%   \includegraphics[width=.8\linewidth]{ROC 1.3.2.png}  
%   \caption{ResNet101 AUC score: 1.0}
%   \label{fig:sub-second}
% \end{subfigure}

% \newline

% \begin{subfigure}{.5\textwidth}
%   \centering
%   % include third image
%   \includegraphics[width=.8\linewidth]{ROC 1.3.3.png}  
%   \caption{VGG19 AUC score: 0.99}
%   \label{fig:sub-third}
% \end{subfigure}
% \begin{subfigure}{.5\textwidth}
%   \centering
%   % include fourth image
%   \includegraphics[width=.8\linewidth]{ROC 1.3.4.png}  
%   \caption{Random Forest AUC score: 0.99}
%   \label{fig:sub-fourth}
% \end{subfigure}
% \paragraph{}
% \caption{Augmented localized ROC curves}
% \label{fig:fig}
% \end{figure}

\begin{figure}[h] % picture
    \centering
    \includegraphics[scale=0.45]{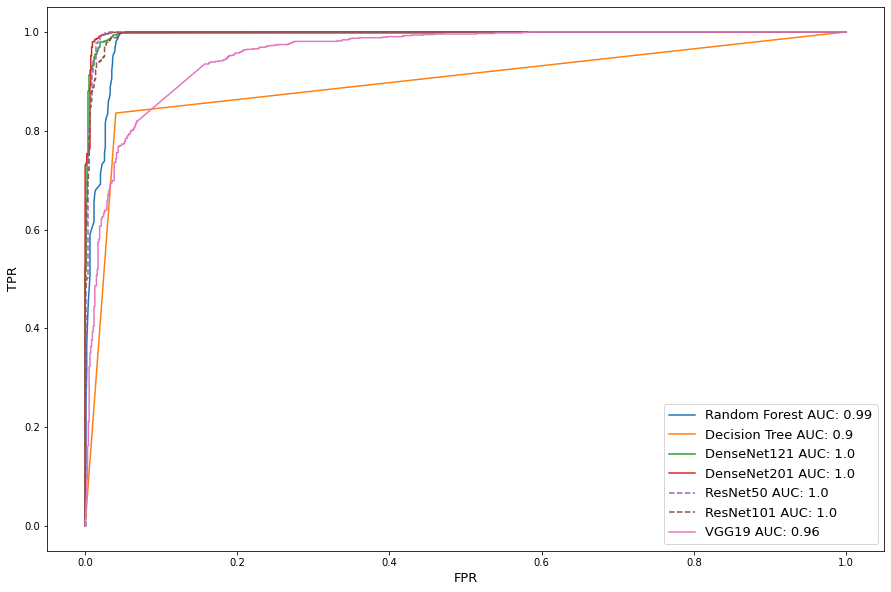}
    \caption{Augmented localized ROC curves}
    \label{figure:figure6}
\end{figure}

% \pagebreak

Table ~\ref{table:table4} shows the classification performance for the augmented data. We see the addition of data allows near perfect score improvement for the deep learning models. However, the increase in sample size decreased overall conventional machine learning algorithm accuracy due to a higher possibility for false predictions. This phenomenon is also reflected in Figure ~\ref{figure:figure6} as deep learning reached AUC scores nearing 1.0, while the Decision Tree AUC score dropped 0.05 from Figure ~\ref{figure:figure5}.

% \pagebreak

We showed it is in fact possible to detect GAN generated cancer scans, even when trained radiologists and their accompanying AI tools fell short in their efforts. In the open trials of the CT-GAN  experiment (\cite{DBLP:journals/corr/abs-1901-03597}), the tumor injected misdiagnosis rate dropped from 99.2\% to 70\%, and though this improvement is drastic in terms of human performed detection, it is far greater than the misdiagnosis rate of deep transfer learning which is <1\%. The monumental limitation of the method presented in this section is the singular focus on FM scans, which allowed for tumor site localization. Providing knowledge on where to look in a 512x512 scan eliminates the additional task of searching for countless possible tampering locations, drastically simplifying the problem. This method of localization cannot be performed on FB scans as the tumor site is unknown in a practical setting.

\subsection{False Benign detection}

The primary focus for this section is classifying FB from untampered scans, where tumors are removed. Unlike the previous section, we do not have the liberty of tumor site localization, i.e.,  the location of the region of interest is unknown in a practical setting.

\subsubsection{Negative Space Reduction}

{As shown in Table ~\ref{table:table2}, machine learning model accuracy is on average approximately 44\% higher than deep learning.
Figure ~\ref{figure:figure3} shows all models have red gradients in the zeroed pixel regions around the scan and the grayed regions surrounding the lungs, indicating the network trained significantly on these areas. Ideally a better neural network shows the maximum gradient class activation at the tumor and nodules, moderate activation inside the lungs, and least activation in the surrounding region. With this in mind,  we apply negative space reduction to remove the surrounding area. }Negative space refers to the zeroed pixels surrounding the ROI. Unlike in Section 3.1.2, the ROI in this section is the entire lung. Each pixel array is sliced in a manner that isolated the lungs and visible nodules. This will preserve the delocalized image without removing any useful image features. 

\begin{figure}[h]%
    \centering
    \subfloat[\centering Before]{{\includegraphics[scale=1]{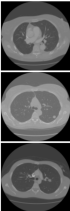} }}%
    \qquad
    \subfloat[\centering After]{{\includegraphics[scale=1]{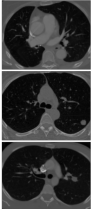} }}%
    \caption{
    Negative space reduced scans \newline
  Row 1: FB samples, 
  Row 2: FM samples, 
  Row 3: Untampered samples}
    \label{fig:figure7}
\end{figure}

\begin{figure}[h] % picture
    \centering
    \includegraphics[scale=0.45]{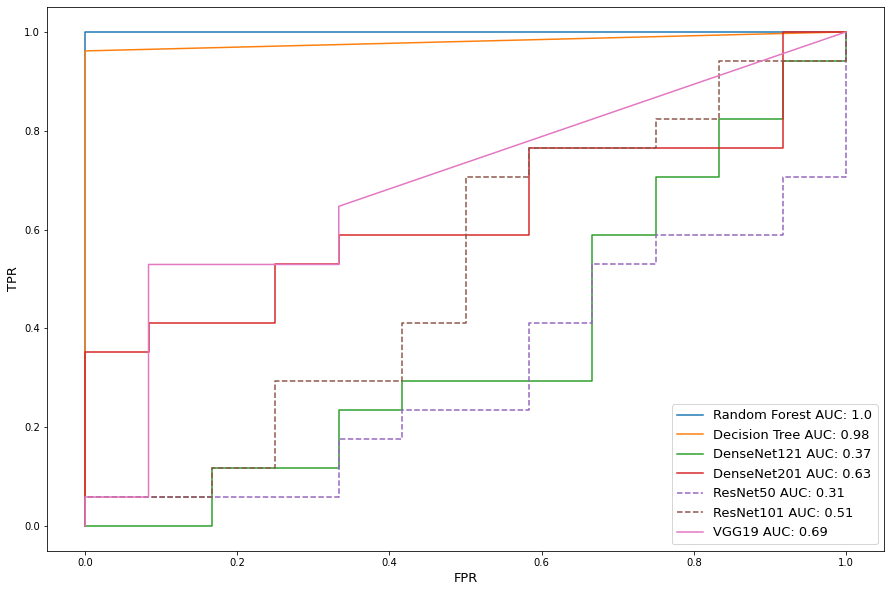}
    \caption{Delocalized image classifier ROC curves}
    \label{figure:figure8}
\end{figure}

Classifier metrics for delocalized image data are presented in Table ~\ref{table:table5}. SVM and Random Forest perform reliably given the small sample size by nature of machine learning algorithms. Certain non-deterministic features are easily extracted without the need for deep learning. Untampered recall is higher than tampered recall for most deep learning models, indicating overcompensated predictions for the untampered class. This overcompensation is suspected to arise due to an unclear decision boundary between the FB and untampered classes: denoised malignant tumor sites are absent in both cases.  AUC scores in Figure ~\ref{figure:figure8} reflect excellent, poor, and failed scores for this task.

% \pagebreak
% \newpage
% \paragraph{}
% \paragraph{}

  \begin{equation}
    Precision=\frac{True\;Positives}{True\;Positives+False\;Positives}
  \end{equation}\break
  \begin{equation}
    Recall=\frac{True\;Positives}{True\;Positives+False\;Negatives}
  \end{equation}

\begin{table}[!htp]\centering
\caption{Delocalized image classifier performance}
\begin{tabular}{lrrrrrr}\toprule
Classifier &Accuracy &untampered Precision &untampered Recall &tampered Precision &tampered Recall \\\midrule
SVM &\textbf{1} &\textbf{1} &\textbf{1} &\textbf{1} &\textbf{1} \\
Random Forest &0.96 &0.9 &\textbf{1} &\textbf{1} &0.938 \\
Decision Tree &0.92 &0.818 &\textbf{1} &\textbf{1} &0.875 \\
DenseNet121 &0.594 &0.467 &0.583 &0.706 &0.6 \\
DenseNet201 &0.5 &0.357 &0.417 &0.611 &0.55 \\
ResNet50 &0.531 &0.421 &0.667 &0.692 &0.45 \\
ResNet101 &0.313 &0.292 &0.583 &0.375 &0.15 \\
VGG19 &0.406 &0.316 &0.5 &0.538 &0.35 \\
\bottomrule
\end{tabular}
\label{table:table5}
\end{table}

% \pagebreak

\newpage
% \pagebreak

% \newpage
% \pagebreak

\subsubsection{Data Augmentation}

\begin{figure}[ht] % picture
    \centering
    \includegraphics[scale=0.45]{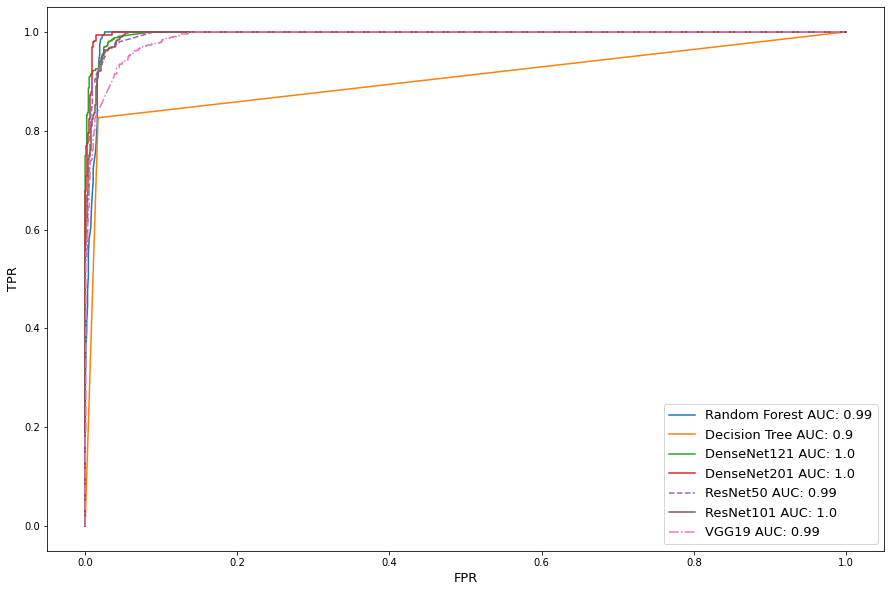}
    \caption{Augmented delocalized ROC curves}
    \label{figure:figure9}
\end{figure}

In this section, we repeat the previous tests using data augmentation. 
In accordance with section 3.1.3, Table ~\ref{table:table6} shows data augmentation of CT scans does improve deep learning accuracy. Random forest performance is on par with ResNet101, showing the  feature pool for 266x340 images could be more representative of the class boundaries when compared to localized 128x128 image slices. Mean accuracy by model, however, is lower by 0.0148\%, equating to 15 additional misdiagnoses per 1000 scan classifications, confirming a larger feature pool is more difficult to learn efficiently. Perfect untampered precision and tampered recall scores, and lower untampered recall and tampered precision on SVM, DenseNet201, and ResNet50 indicate overcompensation on tampered predictions while correctly classifying ~92\% of untampered scans. Given perfect model accuracy is non-viable, this is the preferred scenario considering loss of life is more concerning than expense of resources. AUC scores in Figure ~\ref{figure:figure9} indicate excellent performance.

\begin{table}[h]\centering
\caption{Augmented delocalized image classifier performance}
\begin{tabular}{lrrrrrr}\toprule
Classifier &Accuracy &untampered Precision &untampered Recall &tampered Precision &tampered Recall \\\midrule
SVM &0.959 &\textbf{1} &0.919 &0.925 &\textbf{1} \\
Random Forest &\textbf{0.987} &0.998 &0.977 &0.977 &0.998 \\
Decision Tree &0.913 &0.861 &0.985 &0.983 &0.841 \\
DenseNet121 &0.962 &0.935 &0.993 &0.992 &0.931 \\
DenseNet201 &0.957 &\textbf{1} &0.914 &0.921 &\textbf{1} \\
ResNet101 &0.976 &0.969 &0.984 &0.984 &0.968 \\
ResNet50 &0.955 &\textbf{1} &0.911 &0.918 &\textbf{1} \\
VGG19 &0.933 &0.886 &\textbf{0.995} &\textbf{0.994} &0.872 \\
\bottomrule
\end{tabular}
\paragraph{}
\label{table:table6}
\end{table}

% \newpage

% \begin{figure}[h]
% \begin{subfigure}{.5\textwidth}
%   \centering
%   % include first image
%   \includegraphics[width=.8\linewidth]{ROC 2.2.1.png}  
%   \caption{Random Forest AUC score: 0.99}
%   \label{fig:sub-first}
% \end{subfigure}
% \begin{subfigure}{.5\textwidth}
%   \centering
%   % include second image
%   \includegraphics[width=.8\linewidth]{ROC 2.2.2.png}  
%   \caption{ResNet101 AUC score: 1.0}
%   \label{fig:sub-second}
% \end{subfigure}

% \newline

% \begin{subfigure}{.5\textwidth}
%   \centering
%   % include third image
%   \includegraphics[width=.8\linewidth]{ROC 2.2.3.png}  
%   \caption{DenseNet121 AUC score: 1.0}
%   \label{fig:sub-third}
% \end{subfigure}
% \begin{subfigure}{.5\textwidth}
%   \centering
%   % include fourth image
%   \includegraphics[width=.8\linewidth]{ROC 2.2.4.png}  
%   \caption{ResNet50 AUC score: 0.96}
%   \label{fig:sub-fourth}
% \end{subfigure}
% \paragraph{}
% \caption{Augmented delocalized ROC curves}
% \label{fig:fig}
% \end{figure}

The results of this section indicate the detection of FB scans can be done at high rates of accuracy and recall, accounting also for false negatives. In the open trial conducted by CT-GAN (\cite{DBLP:journals/corr/abs-1901-03597}) researchers, where knowledge of the attack was known, attack success rates dropped a mere 5.8\% from its initial 95.8\%. With a very small margin of improvement arising from educating radiologists on attacks, this work emphasizes the room for improvement in state of the art AI frameworks that assist radiologists in making informed decisions.

% \pagebreak
% \newpage

\subsection{Multiclass Classification}

In this section we test classification of delocalized untampered, FB, and FM images in 3 distinct classes rather than binary. FB and FM images are augmented in order to balance data quantity with the untampered class. Samples are identical to those provided in Figure 6. Multiclass classification results from Table ~\ref{table:table7} show overall untampered precision is significantly higher than FB and FM precision: indicating deep learning is best for drawing decision boundaries between the two classes. However, neither model type can consistently distinguish FM and FB. Most models have higher FM recall metrics than FB recalls while having little deviation between FB metrics. This leads FM to be the favored prediction of the two subtypes.

\begin{table}[!htp]
\caption{Multiclass delocalized image classifier performance }
\small
\small
% \footnotesize
\begin{tabular}{lrrrrrrrr}\toprule
Classifier &Accuracy &untamp. Precision &untamp. Recall &FB Precision &FB Recall &FM Precision &FM Recall \\\midrule
SVM &0.591 &0.998 &0.927 &0.405 &0.407 &0.411 &0.437 \\
Random Forest &0.665 &\textbf{1} &0.923 &0.509 &0.495 &0.52 &0.576 \\
Decision Tree &0.641 &0.769 &0.936 &0.548 &0.45 &0.559 &0.537 \\
DenseNet121 &\textbf{0.804} &0.998 &\textbf{0.978} &\textbf{0.733} &0.655 &\textbf{0.693} &0.78 \\
DenseNet201 &0.777 &0.993 &0.925 &0.687 &0.561 &0.651 &\textbf{0.814} \\
ResNet50 &0.641 &\textbf{1} &0.923 &0.492 &\textbf{1} &0 &0 \\
ResNet101 &0.706 &0.97 &0.974 &0.557 &0.584 &0.592 &0.561 \\
VGG19 &0.657 &0.785 &0.966 &0.595 &0.32 &0.557 &0.687 \\
\bottomrule
\end{tabular}
% \paragraph{}
\label{table:table7}
\end{table}

\section{Conclusion}
This paper tested various machine learning and deep learning models, as well as image isolation methods for GAN based tampering. We showed the detection of artificially generated deformities in medical imaging can be performed with high confidence. Based on the case studies, we found deep learning with localization of the region of interest to be most proficient at classifying tumor injected scans. The same can also be repeated with negative space reduced scans when localization is infeasible. Future works may involve training on hospital scans and integration of real time deep learning based tamper detection into PACS systems.

\section*{CRediT authorship contribution statement}
\textbf{Siddharth Solaiyappan:} Conceptualization, Methodology, Software, Formal analysis, Investigation, Data Curation, Writing - Original Draft.

\textbf{Yuxin Wen:} Methodology, Writing - Review \& Editing, Supervision

\section*{Conflict of Interest Statement}
\paragraph{}The authors have no conflicts of interest to declare.
All co-authors have seen and agree with the contents of the manuscript and
there is no financial interest to report. We certify that the submission is
original work and is not under review at any other publication.

% \newpage
\bibliography{manuscript}

\begin{thebibliography}{}

\bibitem[A.~P.~Reeves, 2011]{nodulereport}
A.~P.~Reeves, A. M.~B. (2011).
\newblock {The Lung Image Database Consortium (LIDC) Nodule Size Report}.

\bibitem[Armato~III et~al., 2015]{lidcdata}
Armato~III, S.~G., McLennan, G., Bidaut, L., McNitt-Gray, M.~F., Meyer, C.~R.,
  Reeves, A.~P., Zhao, B., Aberle, D.~R., Henschke, C.~I., Hoffman, E.~A.,
  Kazerooni, E.~A., MacMahon, H., Van~Beek, E. J.~R., Yankelevitz, D.,
  Biancardi, A.~M., Bland, P.~H., Brown, M.~S., Engelmann, R.~M., Laderach,
  G.~E., Max, D., Pais, R.~C., Qing, D. P.~Y., Roberts, R.~Y., Smith, A.~R.,
  Starkey, A., Batra, P., Caligiuri, P., Farooqi, A., Gladish, G.~W., Jude,
  C.~M., Munden, R.~F., Petkovska, I., Quint, L.~E., Schwartz, L.~H., Sundaram,
  B., Dodd, L.~E., Fenimore, C., Gur, D., Petrick, N., Freymann, J., Kirby, J.,
  Hughes, B., Casteele, A.~V., Gupte, S., Sallam, M., Heath, M.~D., Kuhn,
  M.~H., Dharaiya, E., Burns, R., Fryd, D.~S., Salganicoff, M., Anand, V.,
  Shreter, U., Vastagh, S., Croft, B.~Y., and Clarke, L.~P. (2015).
\newblock {Data From LIDC-IDRI [Data set]. The Cancer Imaging Archive}.

\bibitem[Barragán-Montero et~al., 2021]{BARRAGANMONTERO202152}
Barragán-Montero, A.~M., Thomas, M., Defraene, G., Michiels, S., Haustermans,
  K., Lee, J.~A., and Sterpin, E. (2021).
\newblock Deep learning dose prediction for imrt of esophageal cancer: The
  effect of data quality and quantity on model performance.
\newblock {\em Physica Medica}, 83:52--63.

\bibitem[Beck, 2018]{mcafee}
Beck, C. (2018).
\newblock Mcafee researchers find poor security exposes medical data to
  cybercriminals.

\bibitem[Birajdar and Mankar, 2013]{birajdar2013digital}
Birajdar, G.~K. and Mankar, V.~H. (2013).
\newblock Digital image forgery detection using passive techniques: A survey.
\newblock {\em Digital investigation}, 10(3):226--245.

\bibitem[Chang and Lin, 2011]{chang2011libsvm}
Chang, C.-C. and Lin, C.-J. (2011).
\newblock Libsvm: a library for support vector machines.
\newblock {\em ACM transactions on intelligent systems and technology (TIST)},
  2(3):1--27.

\bibitem[Dua and Graff, 2017]{UCImlrepo}
Dua, D. and Graff, C. (2017).
\newblock {UCI} machine learning repository.

\bibitem[El~Khouli et~al., 2009]{el2009relationship}
El~Khouli, R.~H., Macura, K.~J., Barker, P.~B., Habba, M.~R., Jacobs, M.~A.,
  and Bluemke, D.~A. (2009).
\newblock Relationship of temporal resolution to diagnostic performance for
  dynamic contrast enhanced mri of the breast.
\newblock {\em Journal of Magnetic Resonance Imaging: An Official Journal of
  the International Society for Magnetic Resonance in Medicine},
  30(5):999--1004.

\bibitem[Gite et~al., 2022]{gite2022enhanced}
Gite, S., Mishra, A., and Kotecha, K. (2022).
\newblock Enhanced lung image segmentation using deep learning.
\newblock {\em Neural Computing and Applications}, pages 1--15.

\bibitem[Goodfellow et~al., 2014]{goodfellow2014generative}
Goodfellow, I., Pouget-Abadie, J., Mirza, M., Xu, B., Warde-Farley, D., Ozair,
  S., Courville, A., and Bengio, Y. (2014).
\newblock Generative adversarial nets.
\newblock {\em Advances in neural information processing systems}, 27.

\bibitem[He et~al., 2016]{he2016deep}
He, K., Zhang, X., Ren, S., and Sun, J. (2016).
\newblock Deep residual learning for image recognition.
\newblock In {\em Proceedings of the IEEE conference on computer vision and
  pattern recognition}, pages 770--778.

\bibitem[Huang et~al., 2017]{huang2017densely}
Huang, G., Liu, Z., Van Der~Maaten, L., and Weinberger, K.~Q. (2017).
\newblock Densely connected convolutional networks.
\newblock In {\em Proceedings of the IEEE conference on computer vision and
  pattern recognition}, pages 4700--4708.

\bibitem[Kadam et~al., 2022]{kadam2022efficient}
Kadam, K.~D., Ahirrao, S., and Kotecha, K. (2022).
\newblock Efficient approach towards detection and identification of copy move
  and image splicing forgeries using mask r-cnn with mobilenet v1.
\newblock {\em Computational Intelligence and Neuroscience}, 2022.

\bibitem[Koul et~al., 2022]{koul2022efficient}
Koul, S., Kumar, M., Khurana, S.~S., Mushtaq, F., and Kumar, K. (2022).
\newblock An efficient approach for copy-move image forgery detection using
  convolution neural network.
\newblock {\em Multimedia Tools and Applications}, pages 1--19.

\bibitem[Krishnaraj et~al., 2022]{krishnaraj2022design}
Krishnaraj, N., Sivakumar, B., Kuppusamy, R., Teekaraman, Y., and Thelkar,
  A.~R. (2022).
\newblock Design of automated deep learning-based fusion model for copy-move
  image forgery detection.
\newblock {\em Computational Intelligence and Neuroscience}, 2022.

\bibitem[Liao et~al., 2017]{DBLP:journals/corr/abs-1711-08324}
Liao, F., Liang, M., Li, Z., Hu, X., and Song, S. (2017).
\newblock Evaluate the malignancy of pulmonary nodules using the 3d deep leaky
  noisy-or network.
\newblock {\em CoRR}, abs/1711.08324.

\bibitem[Mirsky et~al., 2019]{DBLP:journals/corr/abs-1901-03597}
Mirsky, Y., Mahler, T., Shelef, I., and Elovici, Y. (2019).
\newblock {CT-GAN:} malicious tampering of 3d medical imagery using deep
  learning.
\newblock {\em CoRR}, abs/1901.03597.

\bibitem[Mohammed et~al., 2021]{Mohammed_2021}
Mohammed, A.~A., Jebur, B.~A., and Younus, K.~M. (2021).
\newblock Hybrid {DCT}-{SVD} based digital watermarking scheme with chaotic
  encryption for medical images.
\newblock {\em {IOP} Conference Series: Materials Science and Engineering},
  1152(1):012025.

\bibitem[Savaridass et~al., 2021]{Pravin_Savaridass_2021}
Savaridass, M.~P., Deepika, R., Aarnika, R., Maniraj, V., Gokilanandhi, P., and
  Kowsika, K. (2021).
\newblock Digital watermarking for medical images using {DWT} and {SVD}
  technique.
\newblock {\em {IOP} Conference Series: Materials Science and Engineering},
  1084(1):012034.

\bibitem[School, 2020]{nctscans}
School, H.~M. (2020).
\newblock Radiation risk from medical imaging.
\newblock {\em Harvard Health Publishing}.

\bibitem[Selvaraju et~al., 2017]{selvaraju2017grad}
Selvaraju, R.~R., Cogswell, M., Das, A., Vedantam, R., Parikh, D., and Batra,
  D. (2017).
\newblock Grad-cam: Visual explanations from deep networks via gradient-based
  localization.
\newblock In {\em Proceedings of the IEEE international conference on computer
  vision}, pages 618--626.

\bibitem[Setio et~al., 2016]{LUNA_DBLP:journals/corr/SetioTBBBC0DFGG16}
Setio, A. A.~A., Traverso, A., de~Bel, T., Berens, M. S.~N., van~den Bogaard,
  C., Cerello, P., Chen, H., Dou, Q., Fantacci, M.~E., Geurts, B., van~der
  Gugten, R., Heng, P., Jansen, B., de~Kaste, M. M.~J., Kotov, V., Lin, J.~Y.,
  Manders, J. T. M.~C., S{'{o}}nora{-}Mengana, A., Garc{'{\i}}a{-}Naranjo,
  J.~C., Prokop, M., Saletta, M., Schaefer{-}Prokop, C., Scholten, E.~T.,
  Scholten, L., Snoeren, M.~M., Torres, E.~L., Vandemeulebroucke, J., Walasek,
  N., Zuidhof, G. C.~A., van Ginneken, B., and Jacobs, C. (2016).
\newblock Validation, comparison, and combination of algorithms for automatic
  detection of pulmonary nodules in computed tomography images: the {LUNA16}
  challenge.
\newblock {\em CoRR}, abs/1612.08012.

\bibitem[Simonyan and Zisserman, 2014]{simonyan2014very}
Simonyan, K. and Zisserman, A. (2014).
\newblock Very deep convolutional networks for large-scale image recognition.
\newblock {\em arXiv preprint arXiv:1409.1556}.

\bibitem[Singh et~al., 2022]{9682744}
Singh, P., Devi, K.~J., Thakkar, H.~K., and Kotecha, K. (2022).
\newblock Region-based hybrid medical image watermarking scheme for robust and
  secured transmission in iomt.
\newblock {\em IEEE Access}, 10:8974--8993.

\bibitem[T. et~al., 2021]{ct}
T., S., K., S., M., U., and K, O. (2021).
\newblock Computed tomography.
\newblock {\em Transparency in Biology. Springer, Singapore}.

\bibitem[Thakur et~al., 2018]{thakur2018blind}
Thakur, T., Singh, K., and Yadav, A. (2018).
\newblock Blind approach for digital image forgery detection.
\newblock {\em International Journal of Computer Applications}, 975:8887.

\end{thebibliography}
\bibliographystyle{apalike}
%\printbibliography

\end{document}